\documentclass[10pt,twocolumn,letterpaper]{article}

\usepackage[pagenumbers]{cvpr} 

\usepackage{graphicx}
\usepackage{amsmath}
\usepackage{amssymb}
\usepackage{booktabs}

\usepackage[pagebackref,breaklinks,colorlinks]{hyperref}

\usepackage[capitalize]{cleveref}
\crefname{section}{Sec.}{Secs.}
\Crefname{section}{Section}{Sections}
\Crefname{table}{Table}{Tables}
\crefname{table}{Tab.}{Tabs.}

\def\cvprPaperID{*****} 
\def\confName{CVPR}
\def\confYear{2023}

\begin{document}

\title{Exploring Expression-related Self-supervised Learning \\ for Affective Behaviour Analysis }

\author{
Fanglei Xue$^{1}$\thanks{Work was down when Fanglei Xue was an intern at Baidu Research.} \; 
Yifan Sun$^{2}$ \;
Yi Yang$^{3}$\thanks{Corresponding author.}\\
$^{1}$ University of Technology Sydney \hspace{1mm} $^{2}$ Baidu Inc. \hspace{1mm} $^{3}$ Zhejiang University \\
{\tt\small xuefanglei19@mails.ucas.ac.cn, sunyf15@tsinghua.org.cn, yangyics@zju.edu.cn}
}

\maketitle

\begin{abstract}
 This paper explores an expression-related self-supervised learning (SSL) method (ContraWarping) to perform expression classification in the 5th Affective Behavior Analysis in-the-wild (ABAW) competition. Affective datasets are expensive to annotate, and SSL methods could learn from large-scale unlabeled data, which is more suitable for this task. By evaluating on the Aff-Wild2 dataset, we demonstrate that ContraWarping outperforms most existing supervised methods and shows great application potential in the affective analysis area. Codes will be released on: \url{https://github.com/youqingxiaozhua/ABAW5}.

\end{abstract}

\section{Introduction} 
Affective computing aims to automatically recognize expressions from static images or videos. With affective computing, people could build applications in society analysis, human-computer interaction systems, driver fatigue monitoring, and so on. For the past few years, many methods~\cite{happy2014automatic,li2017reliable,ruan2021feature,wang2020region,xue2021TransFER,zhang2021RelativeUncertaintya,zhang2022LearnAll} have been proposed to recognize expressions. However, these methods all rely on precise human annotations to learn. Although some of them~\cite{wang2020suppressing,zhang2021RelativeUncertaintya,zhang2022LearnAll} could learn from noisy labels, they can not learn from unlabeled data. Unfortunately, expressions are subjective and subtle, making annotation a large-scale expression database very expensive and limiting the scale of current databases.

Recently, some researchers proposed some self-supervised learning methods to learn from unlabeled data. Contrastive learning-based methods (such as SimCLR~\cite{chen2020SimpleFramework}, MoCo~\cite{He2020MoCo}, BYOL~\cite{grill2020BYOL}, \etc) learn image features from different views of the Siamese network. Differently, MAE~\cite{he2021MaskedAutoencoders} try to reconstruct a masked image to learn semantic features. Some works also adopt these ideas for face tasks. SSPL~\cite{shu2021LearningSpatialSemantic} learns the spatial-semantic relationship of face images by correct rotated patches, face parsing, and area classification tasks. He~\etal~\cite{he2022EnhancingFace} try to benefit the face recognition task by adopting a 3D reconstruction task. TCAE~\cite{li2019TCAE} and FaceCycle~\cite{zhang2021LearningFacial} learn face representation by disentangling pose, expression, and identity features from each other. Most recently, a contrastive learning method ContraWarping~\cite{xue2023unsupervised} is proposed to learn expression-related features by directly simulating muscle movements. All these methods demonstrate their effectiveness in static image databases~\cite{li2017reliable, kollias2021affect, barsoum2016training}.

Aff-wild2~\cite{kollias2023abaw2,kollias2023abaw,kollias2022abaw,kollias2021distribution,kollias2021analysing,kollias2021affect,kollias2020analysing,kollias2019expression,kollias2019face,kollias2019deep,zafeiriou2017aff} is a large-scale video database for ABAW competitions. It annotated 548 videos, around 2.7M frames, into eight pre-defined categories: anger, disgust, fear, happiness, sadness, surprise, neutral, and other. Thanks to the release of this database, we conduct experiments to explore the effectiveness of ContraWarping on this in-the-wild video database. By directly fine-tuning the pre-trained weights from ContraWarping, we get the performance of the validation set of Expression (Expr) Classification Challenge with a Res-50 backbone, significantly outperforming the supervised one.

\begin{figure*}
  \centering
   \includegraphics[width=0.96\linewidth]{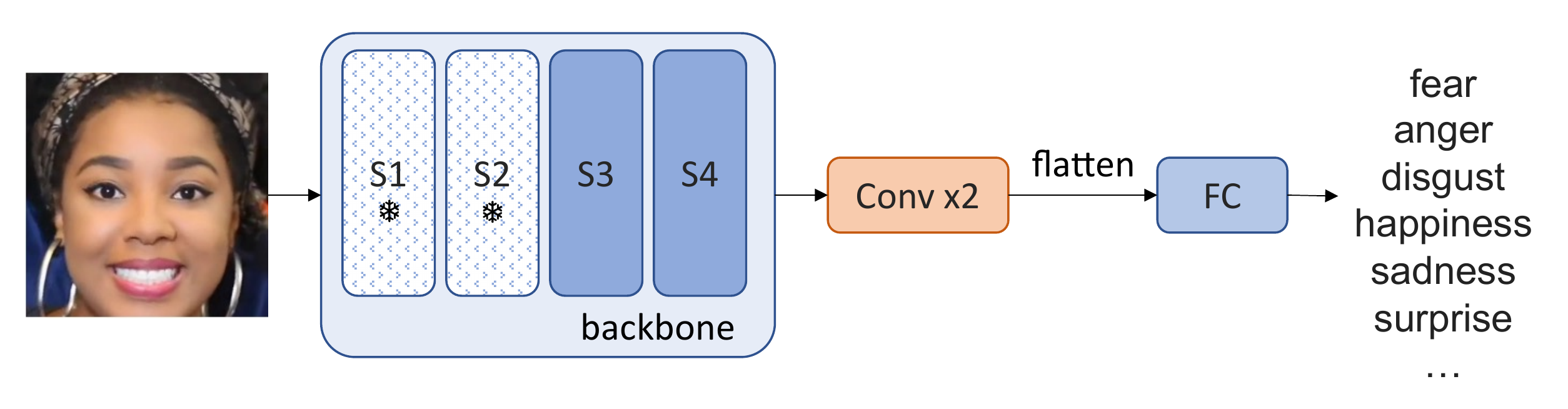}
   \caption{The pipeline of our method. The first two stages of the backbone is freezed. }
   \label{fig:model}
\end{figure*}

\section{Related Works}
Many inspirational methods have been proposed in previous ABAW competitions. We investigate some expression classification methods and multi-task learning methods which, including the Expr task.

Zhang \etal \cite{zhang2023multi} ensemble multiple 2D backbones to extract features for every single frame and concatenate these features to a temporal encoder to explore temporal features. By combining regression layers and classification layers, it learns from multi-task annotation and ranks first in the ABAW4 challenge. It also used MAE pre-trained weights to enhance its performance. Li~\etal~\cite{li2022affective} also use MAE pre-trained weights combined with AffectNet supervised pre-trained weights and ranked 2nd in ABAW4. Zhang~\etal~\cite{zhang2022transformer} proposed a transformer-based fusion module to fuse multi-modality features from audio, image, and word information. Jeong~\etal~\cite{jeong2022facial} extended the DAN model and achieved 2nd in ABAW3. Xue~\etal~\cite{xue2022coarse} utilized a coarse-to-fine cascade network with a temporal smoothing strategy and ranked 3rd in ABAW3. Zhang~\etal~\cite{zhang2021prior} found that AU, VA, and Expr representations are intrinsically associated with each other and proposed a streaming network for multi-task learning.

\section{Method}
Since this paper focuses on exploring the efficiencies of different SSL methods, we adopt a simple framework to perform frame-wise classification. Architecture and implementation details are described below.

\subsection{Architecture}
Following ~\cite{zhang2023multi}, we simply adopt 2D backbones (such ResNet~\cite{he2016deep}, ViT~\cite{dosovitskiy2020image}, \etc) to extract features from every frames. The pipeline of our architecture is illustrated in Fig.~\ref{fig:model}. The face image is proceeded by a backbone to extract feature maps. 
After that, instead of adopting global average pooling, we adopt two convolutional layers with stride 2 to explore spacial relationships and then flatten via the spacial dimension to keep spatial information. One fully-connected (FC) layer is further attached to generate the final classification results.

To evaluate the effectiveness of different pre-trained weights, we freeze the first two stages of the backbone and fine-tune the last two stages as well as the new-added layers.

\subsection{Implementation}
We adopt random cropping and horizontal flip for data augmentation to prevent over-fitting. The model is fine-tuned with the SGD optimizer for 8000 iters. The learning rate is set to 5e-3 with a cosine decay. The batch size is set to 128. Since the adjacent frames in the video are very similar, we randomly sample one frame of every ten frames for training. The average f1 Score across all eight categories on the validation set is reported.

By default, the Res-50~\cite{he2016deep} network without the last classifier is adopted as the backbone. The kernel size of two down-sampling convolution layers is set to 2, and the hidden dimension is 256.

\section{Experiments}

To investigate the effectiveness of ContraWarping on this in-the-wild video database, we conduct experiments with several backbones and pre-trained weights on the validation set of ABAW5. As illustrated in Tab.~\ref{tab:result}, models with more parameters are not always better. APViT~\cite{xue2022vision} is a recently proposed state-of-the-art method that combines both CNN and ViT for feature extraction. It boosts IR-50 from 30.78 to 35.48. However, it fails to outperform Res-50 with ContraWarping pre-trained, which achieves 37.57 on the validation set. The ContraWarping could increase the performance significantly. Even a simple Res-18 could outperform IR-50 with 33.69, indicating that ContraWarping pre-training is more suitable for expression analysis.

\begin{table}
\centering
\begin{tabular}{ccc}
\toprule
Backbone & Pre-trained      & F1-score \\ \midrule
IR-50~\cite{deng2019arcface}    & Sup. MS1M     & 30.78    \\
APViT~\cite{xue2022vision}    & Sup. MS1M     & 35.48   \\
APViT~\cite{xue2022vision}    & Sup. RAF-DB   & 35.63    \\ \midrule
Res-18~\cite{he2016deep}   & ContraWarping & 33.69    \\ 
Res-50~\cite{he2016deep}   & ContraWarping & 37.57   \\
\bottomrule
\end{tabular}
\caption{Results with different backbones and pre-trained weights. Sup. indicates supervised pre-training with manually annotated labels.}
\label{tab:result}
\end{table}

\section{Conclusion}

In this paper, we adopt a simple pipeline to evaluate the effectiveness of ContraWarping, a self-supervised learning method for affective analysis on Aff-Wild2. The ContraWarping could learn expression-related features from unlabeled data by simulating muscle movements. Experiments on Aff-Wild2 indicate that models initialized with ContraWarping pre-trained weights could extract more informative features and performs better than supervised ones.

{\small
\bibliographystyle{ieee_fullname}
\bibliography{egbib}
}

\end{document}